\documentclass{article} % For LaTeX2e
\usepackage{iclr2018_conference,times}
\usepackage{hyperref}
\usepackage{url}
\usepackage{amsthm}
\usepackage{amsmath}
\usepackage{graphicx}
\usepackage{wrapfig}
\usepackage{booktabs}  
\usepackage{subcaption}
\usepackage{lscape}
\usepackage{bm}

\title{Neural Language Modeling \\ by Jointly Learning Syntax and Lexicon}
\author{Yikang Shen, Zhouhan Lin, Chin-Wei Huang \& Aaron Courville 
\\
Department of Computer Science and Operations Research \\
Université de Montréal \\
Montréal, QC H3C3J7, Canada \\
\texttt{\{yi-kang.shen, zhouhan.lin, chin-wei.huang, aaron.courville\}@umontreal.ca} \\
}

\iclrfinalcopy % Uncomment for camera-ready version, but NOT for submission.

\begin{document}
\maketitle

\begin{abstract}
We propose a neural language model capable of unsupervised syntactic structure induction. The model leverages the structure information to form better semantic representations and better language modeling. Standard recurrent neural networks are limited by their structure and fail to efficiently use syntactic information. On the other hand, tree-structured recursive networks usually require additional structural supervision at the cost of human expert annotation. In this paper, We propose a novel neural language model, called the Parsing-Reading-Predict Networks (PRPN), that can simultaneously induce the syntactic structure from unannotated sentences and leverage the inferred structure to learn a better language model. In our model, the gradient can be directly back-propagated from the language model loss into the neural parsing network. Experiments show that the proposed model can discover the underlying syntactic structure and achieve state-of-the-art performance on word/character-level language model tasks.  
\end{abstract}

\section{Introduction}
Linguistic theories generally regard natural language as consisting of two part: a \emph{lexicon}, the complete set of all possible words in a language; and a \emph{syntax}, the set of rules, principles, and processes that govern the structure of sentences \citep{sandra1994morphological}. To generate a proper sentence, tokens are put together with a specific syntactic structure. Understanding a sentence also requires lexical information to provide meanings, and syntactical knowledge to correctly combine meanings. Current neural language models can provide meaningful word represent \citep{bengio2003neural, mikolov2013efficient, chen2013expressive}. However, standard recurrent neural networks only implicitly model syntax, thus fail to efficiently use structure information \citep{tai2015improved}. 
%Aaron: this first paragraph should be about the relation between structure and meaning. 

Developing a deep neural network that can leverage syntactic knowledge to form a better semantic representation has received a great deal of attention in recent years \citep{socher2013recursive, tai2015improved, chung2016hierarchical}. Integrating syntactic structure into a language model is important for different reasons: 1) to obtain a hierarchical representation with increasing levels of abstraction, which is a key feature of deep neural networks and of the human brain \citep{bengio2009learning, lecun2015deep, schmidhuber2015deep}; 2) to capture complex linguistic phenomena, like long-term dependency problem \citep{tai2015improved} and the compositional effects \citep{socher2013recursive}; 3) to provide shortcut for gradient back-propagation \citep{chung2016hierarchical}.

%Aaron: discuss supervised structure induction and it's limitations.
A syntactic parser is the most common source for structure information. Supervised parsers can achieve very high performance on well constructed sentences. Hence, parsers can provide accurate information about how to compose word semantics into sentence semantics \citep{socher2013recursive}, or how to generate the next word given previous words \citep{wu2017sequence}. However, only major languages have treebank data for training parsers, and it request expensive human expert annotation. People also tend to break language rules in many circumstances (such as writing a tweet). These defects limit the generalization capability of supervised parsers.

%Aaron: there is too much "list" stucture. Every paragraph has a list embedded inside. It's not a great style.

Unsupervised syntactic structure induction has been among the longstanding challenges of computational linguistic \citep{klein2002generative, klein2004corpus, bod2006all}. Researchers are interested in this problem for a variety of reasons: to be able to parse languages for which no annotated treebanks exist \citep{marecek2016twelve}; to create a dependency structure to better suit a particular NLP application \citep{wu2017sequence}; to empirically argue for or against the poverty of the stimulus \citep{clark2001unsupervised, chomsky2014aspects}; and to examine cognitive issues in language learning \citep{solan2003automatic}. 

%We make no claims as to the cognitive plausibility of the induction mechanisms we present here. However, our model is designed based on the idea of the tree-structured graph, which could be seen as a specific representational adaptation for language which both funds and limits their competence to acquire specific types of natural languages.
% Yikang: Is this claim make enough sense?

In this paper, we propose a novel neural language model: Parsing-Reading-Predict Networks (PRPN), which can simultaneously induce the syntactic structure from unannotated sentences and leverage the inferred structure to form a better language model.
With our model, we assume that language can be naturally represented as a  tree-structured graph. The model is composed of three parts: 
\begin{enumerate}
\item \textbf{A differentiable neural Parsing Network} uses a convolutional neural network to compute the \emph{syntactic distance}, which represents the syntactic relationships between all successive pairs of words in a sentence, and then makes soft constituent decisions based on the syntactic distance.
\item \textbf{A Reading Network} that recurrently computes an adaptive memory representation to summarize information relevant to the current time step, based on all previous memories that are syntactically and directly related to the current token. 
\item \textbf{A Predict Network} that predicts the next token based on all memories that are syntactically and directly related to the next token. 
\end{enumerate}
We evaluate our model on three tasks: word-level language modeling, character-level language modeling, and unsupervised constituency parsing. The proposed model achieves (or is close to) the state-of-the-art on both word-level and character-level language modeling. The model's unsupervised parsing outperforms some strong baseline models, demonstrating that the structure found by our model is similar to the intrinsic structure provided by human experts. 

%Aaron: the following is redundant and unnecessary.
%The contributions of this paper are:
%\begin{itemize}
%\item We propose a differentiable neural parsing model;
%\item We propose a structured attention mechanism to integrate structure information into sequential recurrent neural networks;
%\item We propose a new Parsing-Reading-Predict framework for language modeling;
%\item The proposed model can induce hierarchical syntactic structures from raw data that is close to ground truth provided by human experts;
%\item The proposed model achieves state-of-the-art performance on both word/character-level language model.
%\end{itemize}

\section{Related Work}
%Aaron: This whole section should be shrunk to about haft of it's current length. Also for each paragraph you need to tie it back to the proposed approach - why are your mentioning it?
The idea of introducing some structures, especially trees, into language 
understanding %Aaron: added this word. to be verified.
to help a downstream task has been explored in various ways. For example, \cite{socher2013recursive, tai2015improved} learn a bottom-up encoder, taking as an input a parse tree supplied from an external parser. There are models that are able to \emph{infer} a tree during test time, while still need supervised signal on tree structure during training. For example, \citep{socher2010learning, alvarez2016tree, zhou2017generative, zhang2015top}, etc. Moreover, \cite{williams2017learning} did an in-depth analysis of recursive models that are able to learn tree structure without being exposed to any grammar trees. Our model is also able to infer tree structure in an unsupervised setting, but different from theirs, it is a recurrent network that implicitly models tree structure through attention.

Apart from the approach of using recursive networks to capture structures, there is another line of research which try to learn recurrent features at multiple scales, which can be dated back to 1990s (e.g. \cite{ElHihi+Bengio-nips8, Schmidhuber91neuralsequence, lin1998learning}). The NARX RNN \citep{lin1998learning} is another example which used a feed forward net taking different inputs with predefined time delays to model long-term dependencies. More recently, \cite{koutnik2014clockwork} also used multiple layers of recurrent networks with different pre-defined updating frequencies. Instead, our model tries to learn the structure from data, rather than predefining it. In that respect, \cite{chung2016hierarchical} relates to our model since it proposes a hierarchical multi-scale structure with binary gates controlling intra-layer connections, and the gating mechanism is learned from data too. The difference is that their gating mechanism controls the updates of higher layers directly, while ours control it softly through an attention mechanism.

% \cite{ElHihi+Bengio-nips8} uses domain specific prior knowledge to create a recurrent network with predefined delays and multiple timescales. 
% Aaron: can we cut this next one? Maybe just add the ref somewhere?
%In similar work, Temporal Kernel RNN (TKRNN) \citep{sutskever2010temporal} has each of the hidden units act as a leaky integrator while keeping the forward and the backward pass efficient. 
%Zhouhan: Cutted.
% Another line of research focuses on models that can learn to exploit multiple time-scales. The approach of \cite{Schmidhuber91neuralsequence, schmidhuber1992learning} learns to chunk sequences recursively in a divide-and-conquer manner. In follow up work, \cite{mozer1992induction}  induces a multi-scale temporal structure. Finally, 

In terms of language modeling, syntactic language modeling can be dated back to \cite{chelba1997structured}. \cite{charniak2001immediate, roark2001probabilistic} have also proposed language models with a top-down parsing mechanism. Recently \cite{dyer2016recurrent, kuncoro2016recurrent} have introduced neural networks into this space. It learns both a discriminative and a generative model with top-down parsing, trained with a supervision signal from parsed sentences in the corpus. There are also dependency-based approaches using neural networks, including \cite{buys2015generative, emami2005neural, titov2010latent}.

Parsers are also related to our work since they are all inferring grammatical tree structure given a sentence. For example, SPINN \citep{bowman2016fast} is a shift-reduce parser that uses an LSTM as its composition function. The transition classifier in SPINN is supervisedly trained on the Stanford PCFG Parser \citep{Klein:2003:AUP:1075096.1075150} output. Unsupervised parsers are more aligned with what our model is doing. \cite{klein2004corpus} presented a generative model for the unsupervised learning of dependency structures. \cite{klein2002generative} is a generative distributional model for the unsupervised induction of natural language syntax which explicitly models constituent yields and contexts. We compare our parsing quality with the aforementioned two papers in Section \ref{unsupervised_parsing_exp}.

\section{Motivation}
\begin{figure}[h]
  \centering
  \includegraphics[width=0.75\linewidth]{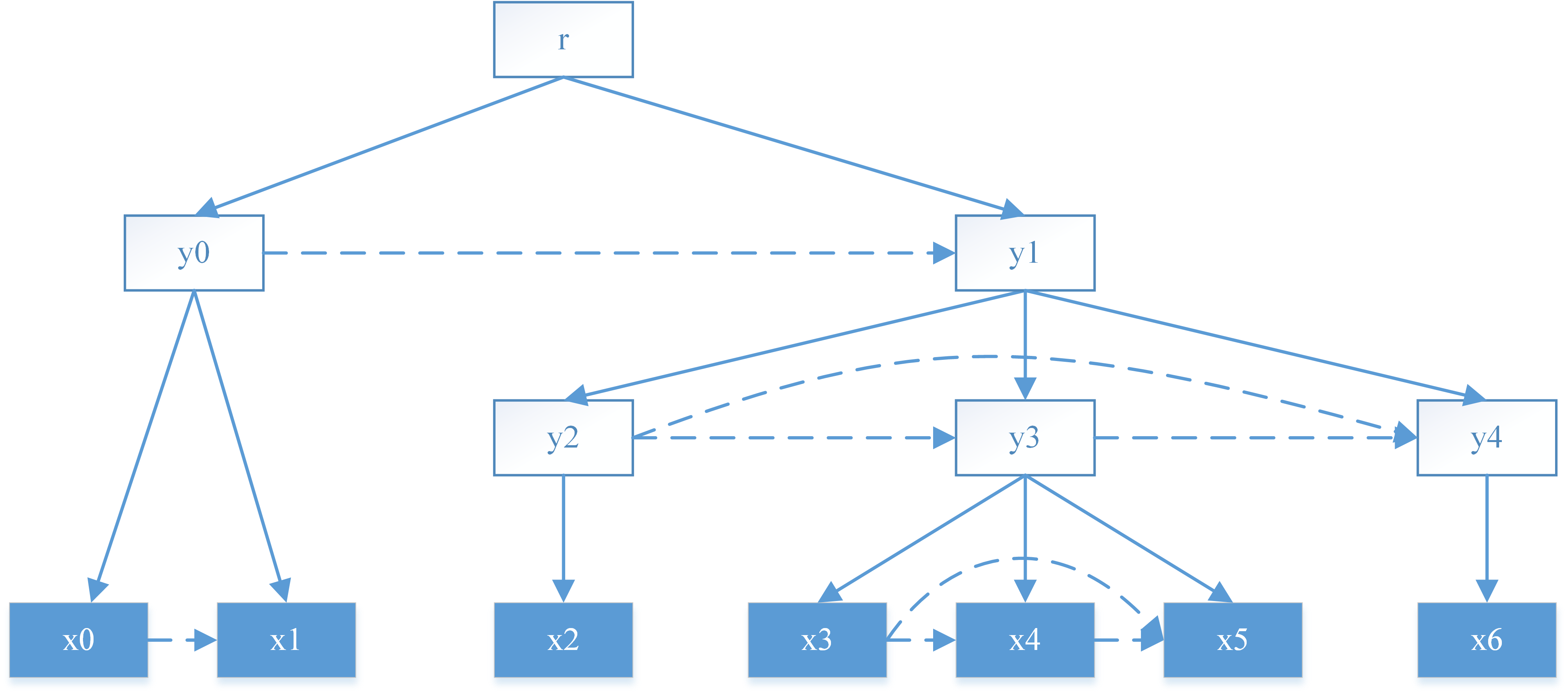}
  \caption{Hard arrow represents syntactic tree structure and parent-to-child dependency relation, dash arrow represents dependency relation between siblings}
  \label{fig_tree}
\end{figure}

Suppose we have a sequence of tokens $x_0,...,x_6$ governed by the tree structure showed in Figure \ref{fig_tree}. The leafs ${x_i}$ are observed tokens. Node $y_i$ represents the meaning of the constituent formed by its leaves $x_{l(y_i)},...,x_{r(y_i)}$, where $l(\cdot)$ and $r(\cdot)$ stands for the leftmost child and right most child. Root $r$ represents the meaning of the whole sequence. Arrows represent the dependency relations between nodes. The underlying assumption is that each node depends only on its parent and its left siblings. 

Directly modeling the tree structure is a challenging task, usually requiring supervision to learn \citep{tai2015improved}. In addition, relying on tree structures can result in a model that is not sufficiently robust to face ungrammatical sentences \citep{hashemi2016evaluation}. 
In contrast, recurrent models provide a convenient way to model sequential data, with the current hidden state only depends on the last hidden state. This makes models more robust when facing nonconforming sequential data, but it suffers from neglecting the real dependency relation that dominates the structure of natural language sentences.
%Aaron: be careful, RNNs do not assume Markov structure over the input. Don't confuse the hidden units with latent variables, this is not the right semantics. 

\begin{figure}[h]
  \centering  
  \includegraphics[width=0.75\linewidth]{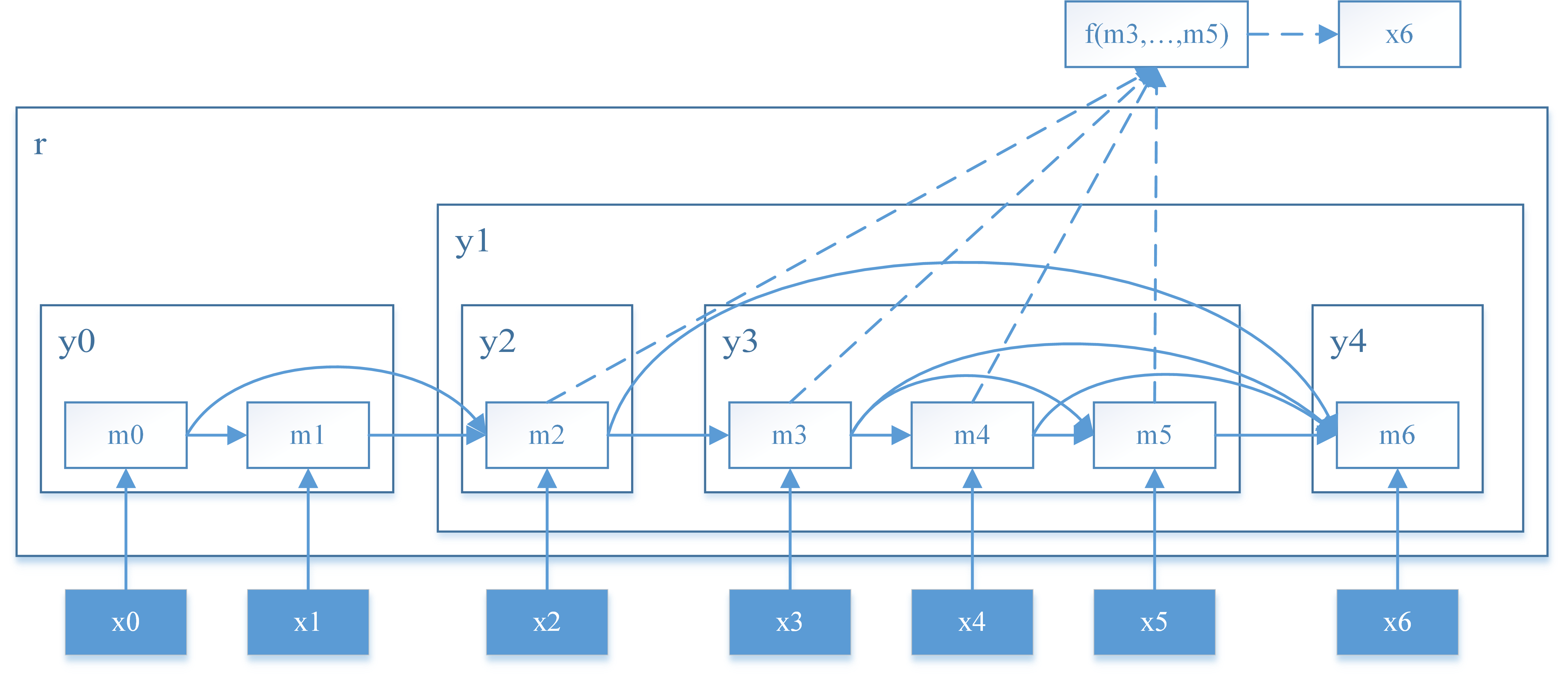}
  \caption{Proposed model architecture, hard line indicate valid connection in Reading Network, dash line indicate valid connection in Predict Network.}
  \label{fig_summ_pred}
\end{figure}

In this paper, we use skip-connection to integrate structured dependency relations with recurrent neural network. In other words, the current hidden state does not only depend on the last hidden state, but also on previous hidden states that have a direct syntactic relation to the current one.
% The \emph{structured attention} masks standard attention in LSTMN with controlling gates, which prevent the model from attending on irrelevant previous hidden states.

Figure \ref{fig_summ_pred} shows the structure of our model. The non-leaf node $y_j$ is represented by a set of hidden states $y_j = \{m_i\}_{l(y_j) \leq i \leq r(y_j)}$, where $l(y_j)$ is the left most descendant leaf and $r(y_j)$ is the right most one. Arrows shows skip connections built by our model according to the latent structure. Skip connections are controlled by gates ${g_i^t}$. In order to define $g_i^t$, we introduce a latent variable $l_t$ to represent local structural context of $x_t$:
\begin{itemize}
\item if $x_t$ is not left most child of any subtree, then $l_t$ is the position of $x_t$'s left most sibling.
\item if $x_t$ is the left most child of a subtree $y_i$, then $l_t$ is the position of the left most child that belongs to the left most sibling of $y_i$. 
\end{itemize}
and gates are defined as:
\begin{equation} \label{eq_gt}
g_i^t = \begin{cases} 1, & l_t \le i < t \\ 0, & 0 < i < l_t \end{cases}
\end{equation}
Given this architecture, the siblings dependency relation is modeled by at least one skip-connect. The skip connection will directly feed information forward, and pass gradient backward. The parent-to-child relation will be implicitly modeled by skip-connect relation between nodes.
%Aaron: I changed the above sentence for language, pls confirm it's still correct.

The model recurrently updates the hidden states according to:
\begin{equation}
m_{t} = h(x_t, m_{0},...,m_{t-1},g_0^t,...,g_{t-1}^t) \label{eq_hidden}
\end{equation}
and the probability distribution for next word is approximated by:
\begin{align}
p(x_{t+1}|x_0,...,x_{t}) &\approx p(x_{t+1};f(m_{0},...,m_t,g_0^{t+1},...,g_{t}^{t+1})) \label{eq_cond}
\end{align}
where ${g_i^t}$ are gates that control skip-connections. Both $f$ and $h$ have a structured attention mechanism that takes $g_i^t$ as input and forces the model to focus on the most related information. Since $l_t$ is an unobserved latent variable, We explain an approximation for ${g_i^t}$ in the next section. The structured attention mechanism is explained in section \ref{sec_reading}.

\section{Modeling Syntactic Structure}
\subsection{Modeling Local Structure} \label{MLS}
In this section we give a probabilistic view on how to model the local structure of language. A detailed elaboration for this section is given in Appendix \ref{appendix_MLS}.
At time step $t$, $p(l_{t}|x_0, ...,x_t)$ represents the probability of choosing one out of $t$ possible local structures. We propose to model the distribution by the Stick-Breaking Process:
\begin{equation}
p(l_{t}=i|x_0, ...,x_t) = (1-\alpha_{i}^{t})\prod_{j=i+1}^{t-1}\alpha_j^{t} 
\label{eq_SB}
\end{equation}
The formula can be understood by noting that after the time step $i+1,...,t-1$ have their probabilities assigned, $\prod_{j=i+1}^{t-1}\alpha_j^{t}$ is remaining probability, $1-\alpha_{i}^{t}$ is the portion of remaining probability that we assign to time step $i$. Variable $\alpha_j^{t}$ is parametrized in the next section.
% * <aaron.courville@gmail.com> 2017-12-20T09:53:26.076Z:
% 
% > $\alpha_j^{t}$
% you need to say more about what alpha represents when you introduce it here.
% 
% ^.
% Yikang: I have added some explanation

% Ideally, at each time step, we can compute the left side of Eq.\ref{eq_hidden} and \ref{eq_cond} by marginalizing over all possible local structures. In other words, we can sampling all possible $l_{t}$, then compute the weighted sum of left side of using $p(l_{t}=i|x_0, ...,x_t)$ as weights.
% % * <aaron.courville@gmail.com> 2017-12-19T21:41:01.976Z:
% % 
% % > Ideally, we can compute equation the expectation of Eq.\ref{eq_hidden} and \ref{eq_cond} by sampling all possible $l_{t}$, then compute the weighted average with $p(l_{t}=i|x_0, ...,x_t)$ as weights.
% % This doesn't make sense. What is meant here?
% % 
% % ^.
% % Yikang: compute eq.2 and 3 actually require marginalizing over all possible unfinished structures
As shown in Appendix \ref{appendix_MLS}, the expectation of gate value $g_i^{t}$ is the Cumulative Distribution Function (CDF) of $p(l_{t}=i|x_0, ...,x_t)$. Thus, we can replace the discrete gate value by its expectation:
\begin{equation} \label{gate_equal_multialpha}
g_{i}^{t} = \mathbf{P}(l_{t}\leq i) = \prod_{j=i+1}^{t-1}\alpha_j^{t}
\end{equation}
With these relaxations, Eq.\ref{eq_hidden} and \ref{eq_cond} can be approximated by using a soft gating vector to update the hidden state and predict the next token. 

\subsection{Parsing Network} \label{sec_tree}
\paragraph{Inferring tree structure with Syntactic Distance}
In Eq.\ref{eq_SB}, $1-\alpha_{j}^{t}$ is the portion of the remaining probability that we assign to position $j$. Because the stick-breaking process should assign high probability to $l_t$, which is the closest constituent-beginning word. The model should assign large $1-\alpha_{j}^{t}$ to words beginning new constituents. While $x_t$ itself is a constituent-beginning word, the model should assign large $1-\alpha_{j}^{t}$ to words beginning larger constituents. In other words, the model will consider longer dependency relations for the first word in constituent. Given the sentence in Figure \ref{fig_tree}, at time step $t=6$, both $1-\alpha_{2}^{6}$ and $1-\alpha_{0}^{6}$ should be close to 1, and all other $1-\alpha_{j}^{6}$ should be close to 0. 
% * <aaron.courville@gmail.com> 2017-12-20T09:59:09.224Z:
% 
% > While $x_t$ itself is a beginning word, the model should assign a higher $1-\alpha_{j}^{t}$ to beginning words of larger constituents. 
% can't parse this.
% 
% ^.
% * <aaron.courville@gmail.com> 2017-12-20T09:56:31.919Z:
% Yikang: the model will look for longer term dependency for first word in a constituent

% > beginning words of constituents
% "words beginning new constituents"?
% 
% ^.

% * <aaron.courville@gmail.com> 2017-12-20T09:54:33.373Z:
% 
% > In Eq.\ref{eq_SB}, $1-\alpha_{j}^{t}$ is the portion of the remaining probability that we assign to position $j$
% I can't parse this at this point. Need more discussion of alpha above to make this make sense.
% 
% ^.

In order to parametrize $\alpha_{j}^t$, our basic hypothesis is that words in the same constituent should have a closer syntactic relation within themselves, and that this syntactical proximity can be represented by a scalar value. From the tree structure point of view, the shortest path between leafs in same subtree is shorter than the one between leafs in different subtree. 

To model syntactical proximity, we introduce a new feature \emph{Syntactic Distance}. For a sentence with length $K$, we define a set of $K$ real valued scalar variables $d_0, ..., d_{K-1}$, with $d_i$ representing a measure of the syntactic relation between the pair of adjacent words $(x_{i-1}, x_{i})$. $x_{-1}$ could be the last word in previous sentence or a padding token. For time step $t$, we want to find the closest words $x_{j}$, that have larger syntactic distance than $d_t$. Thus $\alpha_j^t$ can be defined as:
% * <aaron.courville@gmail.com> 2017-12-20T10:06:59.396Z:
% 
% >  $(x_{i}, x_{i+1})$
% I changed the index here from x_{i-1},x_{i}. so that it makes sense to have a d_0. Please check that this is ok.
% 
% ^ <aaron.courville@gmail.com> 2017-12-20T10:12:40.323Z.
% Yikang: I added a difinition for x_{-1} in the paragraph.
\begin{equation}  \label{soft_alpha}
\alpha_j^t = \frac{\mathrm{hardtanh} \left( (d_t - d_{j}) \cdot \tau \right) + 1}{2} 
\end{equation}
% * <aaron.courville@gmail.com> 2017-12-20T10:08:45.370Z:
% 
% > \begin{equation}  \label{soft_alpha}
% > \alpha_j^t = \frac{\mathrm{hardtanh} \left( (d_t - d_{j+1}) \cdot \tau \right) + 1}{2} 
% > \end{equation}
% why not just use a hard logistic sigmoid if you want it to be between 0,1?
% 
% ^.
% Yikang: this can bring us a consistency of notation with appendix C.
where $\mathrm{hardtanh}(x)=\max(-1, \min(1, x))$. $\tau$ is the temperature parameter that controls the sensitivity of $\alpha_j^t$ to the differences between distances.

The Syntactic Distance has some nice properties that both allow us to infer a tree structure from it and be robust to intermediate non-valid tree structures that the model may encounter during learning. In Appendix \ref{appendix_overlap} and \ref{gd_properties} we list these properties and further explain the meanings of their values.

\begin{figure}[h]
  \centering
  \includegraphics[width=1\linewidth]{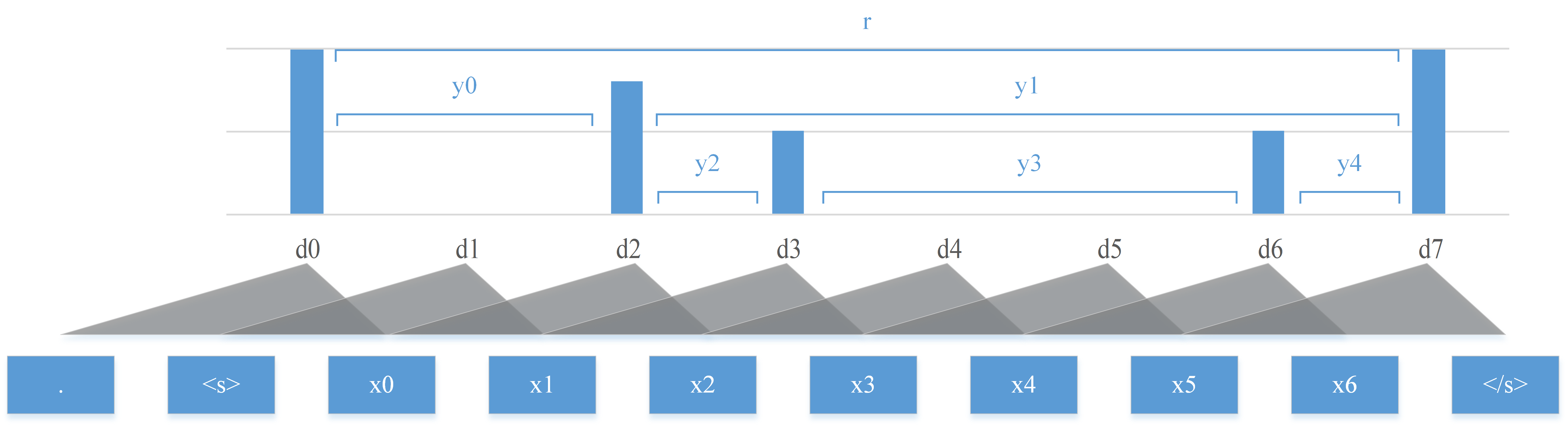}
  \caption{Convolutional network for computing syntactic distance. Gray triangles represent 2 layers of convolution,
$d_0$ to $d_7$ are the syntactic distance output by each of the kernel position. The blue bars indicate the amplitude of $d_i$'s, and $y_i$'s are the inferred constituents.}
% * <aaron.courville@gmail.com> 2017-12-20T10:15:42.129Z:
% 
% > Convolutional network for computing syntactic distance. Gray triangles represent 2 layers of convolution,
% > %convolution kernel (),
% > $d_0$ to $d_7$ are the syntactic distance output by each of the kernel position. The blue bars indicates the amplitude of $d_i$'s, and $y_i$'s are the inferred constituents.
% You need to discuss the y axis (0,0.5,1) - I assume this is alpha, but it should be made explicit.
% 
% ^.
% Yikang: the vertical axis is d_i
  \label{fig_conv}
\end{figure}

\paragraph{Parameterizing Syntactic Distance}
\cite{roark2008classifying} shows that it's possible to identify the beginning and ending words of a constituent using local information. In our model, the syntactic distance between a given token (which is usually represented as a vector word embedding $e_i$) and its previous token $e_{i-1}$, is provided by a convolutional kernel over a set of consecutive previous tokens $e_{i-L}, e_{i-L+1}, ..., e_i$. This convolution is depicted as the gray triangles shown in Figure \ref{fig_conv}. Each triangle here represent 2 layers of convolution. Formally, the syntactic distance $d_i$ between token $e_{i-1}$ and $e_i$ is computed by
\begin{equation} \label{conv_kernel_1}
h_i = \mathrm{ReLU}(W_c \left[ \begin{matrix} e_{i-L} \\ e_{i-L+1} \\ ... \\ e_i \end{matrix} \right] + b_c)
\end{equation}
\begin{equation} \label{conv_kernel_2}
d_i = \mathrm{ReLU} \left(W_d h_i + b_d\right)
\end{equation}
where $W_c$, $b_c$ are the kernel parameters. $W_d$ and $b_d$ can be seen as another convolutional kernel with window size 1, convolved over $h_i$'s. Here the kernel window size $L$ determines how far back into the history node $e_i$ can reach while computing its syntactic distance $d_i$. Thus we call it the \emph{look-back range}. 

Convolving $\bm{h}$ and $\bm{d}$ on the whole sequence with length $K$ yields a set of distances. For the tokens in the beginning of the sequence, we simply pad $L-1$ zero vectors to the front of the sequence in order to get $K-1$ outputs.

%The first convolution is over a set of adjacent tokens before the current token. Formally, given a sequence of word embeddings $E=(\mathbf{e_1}, \mathbf{e_2}, ..., \mathbf{e_i}, ..., \mathbf{e_K})$, where $K$ is the length of the sentence. and the convolutional window size is $L$ for the $i$-th token embedding, the convolution would take $\mathbf{e_{i-L+1}}$ to $\mathbf{e_i}$ as input, and outputs a single vector $\mathbf{H_i}$. We pad $L-1$ zero vectors to the front of the sentence in order to get $K-1$ outputs. 
%The second convolution has window size 1. It takes the $K-1$ vectors ($\mathbf{H_0}$ to $\mathbf{H_{K-1}}$) and map them to K-1 scalar values. We use sigmoid to bound these values to reside between 0 and 1. This values correspond to $d_0, ..., d_7$ in Figure \ref{fig_conv}.

\section{Modeling Language}
\subsection{Reading Network} \label{sec_reading}
The Reading Network generate new states $m_t$ considering on input $x_t$, previous memory states $m_0,...,m_{t-1}$, and gates $g_0^t,...,g_{t-1}^t$, as shown in Eq.\ref{eq_hidden}.

Similar to Long Short-Term Memory-Network (LSTMN) \citep{cheng2016long}, the Reading Network maintains the memory states by maintaining two sets of vectors: a hidden tape $H_{t-1} = \left( h_{t-N_m}, ..., h_{t-1} \right)$, and a memory tape $C_{t-1} = \left( c_{t-L}, ..., c_{t-1} \right)$, where $N_m$ is the upper bound for the memory span. Hidden states $m_i$ is now represented by a tuple of two vectors $(h_i, c_i)$. The Reading Network captures the dependency relation by a modified attention mechanism: \emph{structured attention}. At each step of recurrence, the model summarizes the previous recurrent states via the structured attention mechanism, then performs a normal LSTM update, with hidden and cell states output by the attention mechanism. 

% The recurrent layer then gets updated with the hidden and cell states computed from the structured attention mechanism.
\paragraph{Structured Attention} At each time step $t$, the \textit{read} operation attentively links the current token to previous memories with a structured attention layer:
\begin{align}
% a_i^t &= v^\mathrm{T} \tanh \left( W_h h_i + W_x x_t + W_{\tilde{h}} \tilde{h}_{t-1} \right) \\
k_t &= W_h h_{t-1} + W_x x_t \\
\tilde{s}_i^t &= \mathrm{softmax}(\frac{h_i k_t^{\mathrm{T}}}{\sqrt{\delta_k}})
\end{align}
where, $\delta_k$ is the dimension of the hidden state. 
Modulated by the gates in Eq.\ref{gate_equal_multialpha}, the structured intra-attention weight is defined as:
\begin{align}
s_i^t &= \frac{g_i^t \tilde{s}_i^t}{\sum_i g_i^t}
\end{align}
This yields a probability distribution over the hidden state vectors of previous tokens. 
We can then compute an adaptive summary vector for the previous hidden tape and memory denoting by $\tilde { h } _{ t }$ and $\tilde { c } _{ t }$:
\begin{equation}
\left[ \begin{matrix} \tilde { h } _{ t } \\ \tilde { c } _{ t } \end{matrix} \right] = \sum _{ i=1 }^{ t-1 } s_{ i }^{ t }\cdot m_i = \sum _{ i=1 }^{ t-1 } s_{ i }^{ t }\cdot \left[ \begin{matrix} h_i \\ c_i \end{matrix} \right] 
\end{equation}

Structured attention provides a way to model the dependency relations shown in Figure \ref{fig_tree}.

\paragraph{Recurrent Update} The Reading Network takes $x_t$, $\tilde { c } _{ t }$ and $\tilde { h } _{ t }$ as input, computes the values of $c_t$ and $h_t$ by the LSTM recurrent update \citep{hochreiter1997long}.
Then the \textit{write} operation concatenates $h_t$ and $c_t$ to the end of hidden and memory tape.

\subsection{Predict Network}
% \begin{wrapfigure}{r}{0.4\textwidth}
%   \begin{center}
%     \includegraphics[width=0.38\textwidth]{figure/pred}
%   \end{center}
%   \caption{Predict Network with optional Residual Block (dashed lines)}
%   \label{fig_pred}
% \end{wrapfigure}

Predict Network models the probability distribution of next word $x_{t+1}$, considering on hidden states $m_0,...,m_t$, and gates $g_0^{t+1},...,g_{t}^{t+1}$. Note that, at time step $t$, the model cannot observe $x_{t+1}$ , a temporary estimation of $d_{t+1}$ is computed considering on $x_{t-L},...,x_t$:
\begin{equation} \label{eq_g_estimate}
	d'_{t+1}=\mathrm{ReLU}(W'_d h_t + b'_d)
\end{equation}
From there we compute its corresponding $\{\alpha^{t+1}\}$ and $\{g_i^{t+1}\}$ for Eq.\ref{eq_cond}.
We parametrize $f(\cdot)$ function as:
\begin{align}
f(m_{0},...,m_t,g_0^{t+1},...,g_{t}^{t+1}) &= \hat{f}([h_{l:t-1},h_{t}]) \label{eq_pred}
\end{align}
where $h_{l:t-1}$ is an adaptive summary of $h_{l_{t+1} \leq i \leq t-1}$, output by structured attention controlled by $g_0^{t+1},...,g_{t-1}^{t+1}$. $\hat{f}(\cdot)$ could be a simple feed-forward MLP, or more complex architecture, like ResNet, to add more depth to the model. 

% \section{Training Details}
% The model is end-to-end trained on language modeling dataset. When training word level language models, we use a batch size of 64, truncated back-propagation with 35 timesteps, and we feed final memory from the previous batch as the initial memory of next one. At the beginning of training and test time, the model start with zero. Input and output embedding are shared, as in \cite{} and \cite{}.

% Optimization is performed by Adam with learning rate $lr = 0.003$, weight decay $w_{decay} = 10^{-6}$, $\beta_1 = 0$, $\beta_2 = 0.999$ and $\sigma = 10^{-9}$. The learning rate is multiplied by 0.1 whenever validation performance does not improve ever during 2 checkpoints. These checkpoints are performed at the end of each epoch.

% For character level models, batch size is 64, truncated back-propagation is performed every 100 timesteps. Adams parameters are $\beta_1 = 0.9$, $\beta_2 = 0.999$ and $\sigma = 10^{-8}$. Char embeddings are not shared.

\section{Experiments}
We evaluate the proposed model on three tasks, character-level language modeling, word-level language modeling, and unsupervised constituency parsing. 
% Language modeling task aims at learning the probability over sequences by minimizing the negative log-likelihood of the training sequences:
% \begin{equation}
% \min_{\theta} - \frac{1}{N} \sum_{n=1}^N \sum_{t=1}^{T^n} log p(x_t^n | x_{<t}^n; \theta)
% \end{equation}
% where $\theta$ is the model

\subsection{Character-level Language Model}
From a character-level view, natural language is a discrete sequence of data, where discrete symbols form a distinct and shallow tree structure: the sentence is the root, words are children of the root, and characters are leafs. However, compared to word-level language modeling, character-level language modeling requires the model to handle longer-term dependencies. We evaluate a character-level variant of our proposed language model over a preprocessed version of the Penn Treebank (PTB) and Text8 datasets.

When training, we use truncated back-propagation, and feed the final memory position from the previous batch as the initial memory of next one. At the beginning of training and test time, the model initial hidden states are filled with zero. Optimization is performed with Adam using learning rate $lr = 0.003$, weight decay $w_{decay} = 10^{-6}$, $\beta_1 = 0.9$, $\beta_2 = 0.999$ and $\sigma = 10^{-8}$. We carry out gradient clipping with maximum norm 1.0.	 The learning rate is multiplied by 0.1 whenever validation performance does not improve during 2 checkpoints. These checkpoints are performed at the end of each epoch. We also apply layer normalization \citep{ba2016layer} to the Reading Network and batch normalization to the Predict Network and parsing network. For all of the character-level language modeling experiments, we apply the same procedure, varying only the number of hidden units, mini-batch size and dropout rate. 

\paragraph{Penn Treebank} we process the Penn Treebank dataset \citep{marcus1993building} by following the procedure introduced in \citep{mikolov2012subword}. For character-level PTB, Reading Network has two recurrent layers, Predict Network has one residual block. Hidden state size is 1024 units. The input and output embedding size are 128, and not shared. Look-back range $L=10$, temperature parameter $\tau = 10$, upper band of memory span $N_m=20$. We use a batch size of 64, truncated back-propagation with 100 timesteps. The values used of dropout on input/output embeddings, between recurrent layers, and on recurrent states were (0, 0.25, 0.1) respectively.

\begin{figure*}
    \centering
    \setlength{\leftskip}{-110pt}
    \begin{subfigure}[b]{1.5\textwidth}
        \includegraphics[width=\textwidth]{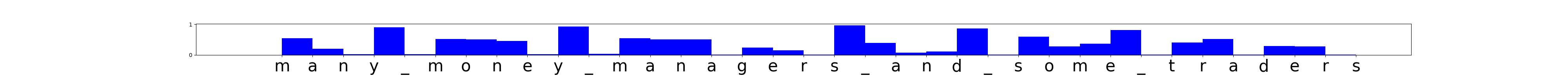}
    \end{subfigure}
    \begin{subfigure}[b]{1.5\textwidth}
        \includegraphics[width=\textwidth]{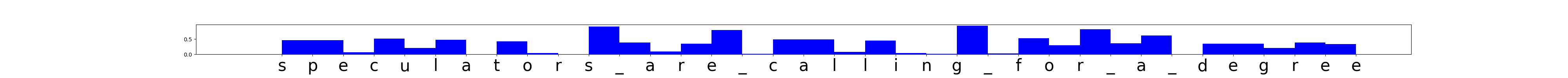}
    \end{subfigure}
    \begin{subfigure}[b]{1.5\textwidth}
        \includegraphics[width=\textwidth]{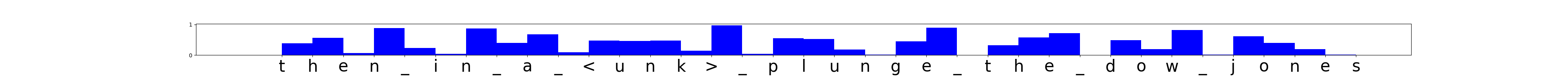}
    \end{subfigure}
    \caption{Syntactic distance estimated by Parsing Network. The model is trained on PTB dataset at the character level. Each blue bar is positioned between two characters, and represents the syntactic distance between them. From these distances we can infer a tree structure according to Section \ref{sec_tree}. }
    \label{fig_char}
\end{figure*}

\begin{table}[ht]  
\centering  
  \begin{tabular}{ c c }
    \toprule[2pt]
    Model & BPC \\
    \hline
    Norm-stabilized RNN \citep{krueger2015regularizing} &  1.48 \\
    CW-RNN \citep{koutnik2014clockwork} & 1.46 \\
    HF-MRNN \citep{mikolov2012subword} & 1.41 \\
    MI-RNN \citep{wu2016multiplicative} & 1.39 \\
    ME n-gram \citep{mikolov2012subword} & 1.37 \\
    BatchNorm LSTM \citep{cooijmans2016recurrent} & 1.32 \\
    Zoneout RNN \citep{krueger2016zoneout} & 1.27 \\
    HyperNetworks \citep{ha2016hypernetworks} & 1.27 \\
    LayerNorm HM-LSTM \citep{chung2016hierarchical} & 1.24 \\
    LayerNorm HyperNetworks \citep{ha2016hypernetworks} & 1.23 \\
    \hline
    PRPN & \textbf{1.202} \\
    \toprule[2pt]
  \end{tabular}
  \caption{BPC on the Penn Treebank test set}
  \label{tb_char_ptb}
\end{table}

In Figure \ref{fig_char}, we visualize the syntactic distance estimated by the Parsing Network, while reading three different sequences from the PTB test set. We observe that the syntactic distance tends to be higher between the last character of a word and a space, which is a reasonable breakpoint to separate between words. In other words, if the model sees a space, it will attend on all previous step. If the model sees a letter, it will attend no further then the last space step. The model autonomously discovered to avoid inter-word attention connection, and use the hidden states of space (separator) tokens to summarize previous information. This is strong proof that the model can understand the latent structure of data. As a result our model achieve state-of-the-art performance and significantly outperform baseline models. It is worth noting that HM-LSTM \citep{chung2016hierarchical} also unsupervisedly induce similar structure from data. But discrete operations in HM-LSTM make their training procedure more complicated then ours.

% We haven't finish training char-level text8 model yet
% \paragraph{Text8} \citep{mahoney2011large} consists of 100M characters extracted from the Wikipedia corpus, all punctuations and special characters are eliminated, and all characters are in lower case. We follow the data splits used in \cite{mikolov2012subword}. For character-level \texttt{text8}, we largely follow our settings for PTB, except that embedding size is stt to 800 and hidden states 1200 instead. We truncate the BPTT with 160b timesteps, and the values used for dropout on input/output embeddings, between Recurrent Layers, on recurrent states, and attention dropout were (0.1, 0.15, 0.1, 0.05), respectively.

% \begin{table}[ht]  
% \centering  
%   \begin{tabular}{ c c }
%     \toprule[2pt]
%     Model & BPC \\
%     \hline
%     td-LSTM (Zhang et al., 2016) & 1.63 \\
%     HF-MRNN (Mikolov et al., 2012) & 1.54 \\
%     MI-RNN (Wu et al., 2016) & 1.52 \\
%     Skipping-RNN (Pachitariu \& Sahani, 2013) & 1.48 \\
%     MI-LSTM (Wu et al., 2016) & 1.44 \\
%     BatchNorm LSTM (Cooijmans et al., 2016) & 1.36 \\
%     \hline
%     Ours & 1.44 \\
%     \toprule[2pt]
%   \end{tabular}
%   \caption{PPL on the Penn Treebank test set}
% \end{table}

\subsection{Word-level Language Model}
Comparing to character-level language modeling, word-level language modeling needs to deal with complex syntactic structure and various linguistic phenomena. But it has less long-term dependencies. We evaluate the word-level variant of our language model on a preprocessed version of the Penn Treebank (PTB) \citep{marcus1993building} and Text8 \citep{mahoney2011large} dataset.

We apply the same procedure and hyper-parameters as in character-level language model. Except optimization is performed with Adam with $\beta_1 = 0$. This turns off the exponential moving average for estimates of the means of the gradients \citep{melis2017state}. We also adapt the number of hidden units, mini-batch size and the dropout rate according to the different tasks. 

\paragraph{Penn Treebank} we process the Penn Treebank dataset \citep{mikolov2012subword} by following the procedure introduced in \citep{mikolov2010recurrent}. For word-level PTB, the Reading Network has two recurrent layers and the Predict Network do not have residual block. The hidden state size is 1200 units and the input and output embedding sizes are 800, and shared \citep{inan2016tying, E17-2025}. Look-back range $L=5$, temperature parameter $\tau = 10$ and the upper band of memory span $N_m=15$. We use a batch size of 64, truncated back-propagation with 35 time-steps. The values used of dropout on input/output embeddings, between recurrent layers, and on recurrent states were (0.7, 0.5, 0.5) respectively.

\begin{table}[h]  
\centering  
  \begin{tabular}{ c c }
    \toprule[2pt]
    Model & PPL \\
    \hline
    RNN-LDA + KN-5 + cache \citep{mikolov2012context} &  92.0 \\
    LSTM \citep{zaremba2014recurrent} & 78.4 \\
    Variational LSTM \citep{kim2016character} & 78.9 \\
    CharCNN \citep{kim2016character} & 78.9 \\
    Pointer Sentinel-LSTM \citep{merity2016pointer} & 70.9 \\
    LSTM + continuous cache pointer \citep{grave2016improving} & 72.1 \\
    Variational LSTM (tied) + augmented loss \citep{inan2016tying} & 68.5 \\
    Variational RHN (tied) \citep{zilly2016recurrent} & 65.4 \\
    NAS Cell (tied)  \citep{zoph2016neural} & 62.4 \\
    4-layer skip connection LSTM (tied) \citep{melis2017state} & \textbf{58.3} \\
%     3-layer AWD-LSTM (tied) \citep{merity2017regularizing} & \textbf{57.3} \\
    \hline
    PRPN & 61.98 \\
    \toprule[2pt]
  \end{tabular}
  \caption{PPL on the Penn Treebank test set}
  \label{tab_ptb_word}
\end{table}

\begin{table}[h]  
\centering  
  \begin{tabular}{ l c }
    \toprule[2pt]
    Model & PPL \\
    \hline
    PRPN & 61.98 \\
     - Parsing Net & 64.42 \\
     - Reading Net Attention & 64.63 \\
     - Predict Net Attention & 63.65 \\
    Our 2-layer LSTM & 65.81 \\
    \toprule[2pt]
  \end{tabular}
  \caption{Ablation test on the Penn Treebank. ``- Parsing Net'' means that we remove Parsing Network and replace Structured Attention with normal attention mechanism; ``- Reading Net Attention'' means that we remove Structured Attention from Reading Network, that is equivalent to replace Reading Network with a normal 2-layer LSTM; ``- Predict Net Attention'' means that we remove Structured Attention from Predict Network, that is equivalent to have a standard projection layer; ``Our 2-layer LSTM'' is equivalent to remove Parsing Network and remove Structured Attention from both Reading and Predict Network.}
  \label{tab_ablation}
\end{table}

\paragraph{Text8} dataset contains 17M training tokens and has a vocabulary size of 44k words. The dataset is partitioned into a training set (first 99M characters) and a development set (last 1M characters) that is used to report performance. As this dataset contains various articles from Wikipedia, the longer term information (such as current topic) plays a bigger role than in the PTB experiments \citep{mikolov2014learning}. We apply the same procedure and hyper-parameters as in character-level PTB, except we use a batch size of 128. The values used of dropout on input/output embeddings, between Recurrent Layers and on recurrent states were (0.4, 0.2, 0.2) respectively.

\begin{table}[h]  
\centering  
  \begin{tabular}{ c c }
    \toprule[2pt]
    Model & PPL \\
    \hline
    LSTM-500 \citep{mikolov2014learning} &  156 \\
    SCRNN \citep{mikolov2014learning} & 161 \\
    MemNN \citep{sukhbaatar2015end} & 147 \\
    LSTM-1024 \citep{grave2016improving} & 121 \\
    LSTM + continuous cache pointer \citep{grave2016improving} & 99.9 \\
    \hline
    PRPN & \textbf{81.64} \\
    \toprule[2pt]
  \end{tabular}
  \caption{PPL on the Text8 valid set}
  \label{tab_text8_word}
\end{table}

In Table \ref{tab_ptb_word}, our results are comparable to the state-of-the-art methods. Since we do not have the same computational resource used in \citep{melis2017state} to tune hyper-parameters at large scale, we expect that our model could achieve better performance after an aggressive hyperparameter tuning process.
As shown in Table \ref{tab_text8_word}, our method outperform baseline methods. It is worth noticing that the \emph{continuous cache pointer} can also be applied to output of our Predict Network without modification. Visualizations of tree structure generated from learned PTB language model are included in Appendix \ref{appendix_tree_fig}.
In Table \ref{tab_ablation}, we show the value of test perplexity for different variants of PRPN, each variant remove part of the model. By removing Parsing Network, we observe a significant drop of performance. This stands as empirical evidence regarding the benefit of having structure information to control attention.

\subsection{Unsupervised Constituency Parsing}  \label{unsupervised_parsing_exp}
The unsupervised constituency parsing task compares hte tree structure inferred by the model with those annotated by human experts. The experiment is performed on WSJ10 dataset. WSJ10 is the 7422 sentences in the Penn Treebank Wall Street Journal section which contained 10 words or less after the removal of punctuation and null elements. Evaluation was done by seeing whether proposed constituent spans are also in the Treebank parse, measuring unlabeled F1 ($\mathrm{UF}_1$) of unlabeled constituent precision and recall. Constituents which could not be gotten wrong (those of span one and those spanning entire sentences) were discarded. Given the mechanism discussed in Section \ref{sec_tree}, our model generates a binary tree.
%, which is a common problem faced by previous unsupervised constituency parsing model \citep{klein2002generative,bod2006all}.
Although standard constituency parsing tree is not limited to binary tree. Previous unsupervised constituency parsing model also generate binary trees \citep{klein2002generative,bod2006all}. Our model is compared with the several baseline methods, that are explained in Appendix \ref{appendix_parser_baseline}.

Different from the previous experiment setting, the model treat each sentence independently during train and test time. When training, we feed one batch of sentences at each iteration. In a batch, shorter sentences are padded with 0. At the beginning of the iteration, the model's initial hidden states are filled with zero. When testing, we feed on sentence one by one to the model, then use the gate value output by the model to recursively combine tokens into constituents, as described in Appendix \ref{appendix_tree_fig}. 

\begin{table}[h]  
\centering  
  \begin{tabular}{ c c }
    \toprule[2pt]
    Model & $\mathrm{UF}_1$ \\
    \hline
    LBRANCH &  28.7 \\
    RANDOM & 34.7 \\
    DEP-PCFG \citep{carroll1992two} & 48.2 \\
    RBRANCH & 61.7 \\
    CCM \citep{klein2002generative} & 71.9 \\
    DMV+CCM \citep{klein2005natural} & 77.6 \\
    UML-DOP \citep{bod2006all} & \textbf{82.9} \\
    \hline
    PRPN & 70.02 \\
    \hline
    UPPER BOUND & 88.1 \\
    \toprule[2pt]
  \end{tabular}
  \caption{Parsing Performance on the WSJ10 dataset}
  \label{tb_parser}
\end{table}

Table \ref{tb_parser} summarizes the results. Our model significantly outperform the RANDOM baseline indicate a high consistency with human annotation. Our model also shows a comparable performance with CCM model. In fact our parsing network and CCM both focus on the relation between successive tokens.  As described in Section \ref{sec_tree}, our model computes syntactic distance between all successive pair of tokens, then our parsing algorithm recursively assemble tokens into constituents according to the learned distance. CCM also recursively model the probability whether a contiguous subsequences of a sentence is a constituent. Thus, one can understand how our model is outperformed by DMV+CCM and UML-DOP models. The DMV+CCM model has extra information from a dependency parser. The UML-DOP approach captures both contiguous and non-contiguous lexical dependencies \citep{bod2006all}.

\section{Conclusion}
In this paper, we propose a novel neural language model that can simultaneously induce the syntactic structure from unannotated sentences and leverage the inferred structure to learn a better language model. We introduce a new neural parsing network: Parsing-Reading-Predict Network, that can make differentiable parsing decisions. We use a new structured attention mechanism to control skip connections in a recurrent neural network. Hence induced syntactic structure information can be used to improve the model's performance. Via this mechanism, the gradient can be directly back-propagated from the language model loss function into the neural Parsing Network. The proposed model achieve (or is close to) the state-of-the-art on both word/character-level language modeling tasks. Experiment also shows that the inferred syntactic structure highly correlated to human expert annotation.

\section*{Acknowledgement}
The authors would like to thank Timothy J. O'Donnell and Chris Dyer for the helpful discussions.

\bibliography{ref}

\begin{thebibliography}{65}
\providecommand{\natexlab}[1]{#1}
\providecommand{\url}[1]{\texttt{#1}}
\expandafter\ifx\csname urlstyle\endcsname\relax
  \providecommand{\doi}[1]{doi: #1}\else
  \providecommand{\doi}{doi: \begingroup \urlstyle{rm}\Url}\fi

\bibitem[Alvarez-Melis \& Jaakkola(2016)Alvarez-Melis and
  Jaakkola]{alvarez2016tree}
David Alvarez-Melis and Tommi~S Jaakkola.
\newblock Tree-structured decoding with doubly-recurrent neural networks.
\newblock 2016.

\bibitem[Ba et~al.(2016)Ba, Kiros, and Hinton]{ba2016layer}
Jimmy~Lei Ba, Jamie~Ryan Kiros, and Geoffrey~E Hinton.
\newblock Layer normalization.
\newblock \emph{arXiv preprint arXiv:1607.06450}, 2016.

\bibitem[Bengio et~al.(2003)Bengio, Ducharme, Vincent, and
  Jauvin]{bengio2003neural}
Yoshua Bengio, R{\'e}jean Ducharme, Pascal Vincent, and Christian Jauvin.
\newblock A neural probabilistic language model.
\newblock \emph{Journal of machine learning research}, 3\penalty0
  (Feb):\penalty0 1137--1155, 2003.

\bibitem[Bengio et~al.(2009)]{bengio2009learning}
Yoshua Bengio et~al.
\newblock Learning deep architectures for ai.
\newblock \emph{Foundations and trends{\textregistered} in Machine Learning},
  2\penalty0 (1):\penalty0 1--127, 2009.

\bibitem[Bod(2006)]{bod2006all}
Rens Bod.
\newblock An all-subtrees approach to unsupervised parsing.
\newblock In \emph{Proceedings of the 21st International Conference on
  Computational Linguistics and the 44th annual meeting of the Association for
  Computational Linguistics}, pp.\  865--872. Association for Computational
  Linguistics, 2006.

\bibitem[Bowman et~al.(2016)Bowman, Gauthier, Rastogi, Gupta, Manning, and
  Potts]{bowman2016fast}
Samuel~R Bowman, Jon Gauthier, Abhinav Rastogi, Raghav Gupta, Christopher~D
  Manning, and Christopher Potts.
\newblock A fast unified model for parsing and sentence understanding.
\newblock \emph{arXiv preprint arXiv:1603.06021}, 2016.

\bibitem[Buys \& Blunsom(2015)Buys and Blunsom]{buys2015generative}
Jan Buys and Phil Blunsom.
\newblock Generative incremental dependency parsing with neural networks.
\newblock In \emph{Proceedings of the 53rd Annual Meeting of the Association
  for Computational Linguistics and the 7th International Joint Conference on
  Natural Language Processing (Volume 2: Short Papers)}, volume~2, pp.\
  863--869, 2015.

\bibitem[Carroll \& Charniak(1992)Carroll and Charniak]{carroll1992two}
Glenn Carroll and Eugene Charniak.
\newblock \emph{Two experiments on learning probabilistic dependency grammars
  from corpora}.
\newblock Department of Computer Science, Univ., 1992.

\bibitem[Charniak(2001)]{charniak2001immediate}
Eugene Charniak.
\newblock Immediate-head parsing for language models.
\newblock In \emph{Proceedings of the 39th Annual Meeting on Association for
  Computational Linguistics}, pp.\  124--131. Association for Computational
  Linguistics, 2001.

\bibitem[Chelba(1997)]{chelba1997structured}
Ciprian Chelba.
\newblock A structured language model.
\newblock In \emph{Proceedings of the eighth conference on European chapter of
  the Association for Computational Linguistics}, pp.\  498--500. Association
  for Computational Linguistics, 1997.

\bibitem[Chen et~al.(2013)Chen, Perozzi, Al-Rfou, and
  Skiena]{chen2013expressive}
Yanqing Chen, Bryan Perozzi, Rami Al-Rfou, and Steven Skiena.
\newblock The expressive power of word embeddings.
\newblock \emph{arXiv preprint arXiv:1301.3226}, 2013.

\bibitem[Cheng et~al.(2016)Cheng, Dong, and Lapata]{cheng2016long}
Jianpeng Cheng, Li~Dong, and Mirella Lapata.
\newblock Long short-term memory-networks for machine reading.
\newblock \emph{arXiv preprint arXiv:1601.06733}, 2016.

\bibitem[Chomsky(2014)]{chomsky2014aspects}
Noam Chomsky.
\newblock \emph{Aspects of the Theory of Syntax}, volume~11.
\newblock MIT press, 2014.

\bibitem[Chung et~al.(2016)Chung, Ahn, and Bengio]{chung2016hierarchical}
Junyoung Chung, Sungjin Ahn, and Yoshua Bengio.
\newblock Hierarchical multiscale recurrent neural networks.
\newblock \emph{arXiv preprint arXiv:1609.01704}, 2016.

\bibitem[Clark(2001)]{clark2001unsupervised}
Alexander Clark.
\newblock Unsupervised induction of stochastic context-free grammars using
  distributional clustering.
\newblock In \emph{Proceedings of the 2001 workshop on Computational Natural
  Language Learning-Volume 7}, pp.\ ~13. Association for Computational
  Linguistics, 2001.

\bibitem[Cooijmans et~al.(2016)Cooijmans, Ballas, Laurent, G{\"u}l{\c{c}}ehre,
  and Courville]{cooijmans2016recurrent}
Tim Cooijmans, Nicolas Ballas, C{\'e}sar Laurent, {\c{C}}a{\u{g}}lar
  G{\"u}l{\c{c}}ehre, and Aaron Courville.
\newblock Recurrent batch normalization.
\newblock \emph{arXiv preprint arXiv:1603.09025}, 2016.

\bibitem[Dyer et~al.(2016)Dyer, Kuncoro, Ballesteros, and
  Smith]{dyer2016recurrent}
Chris Dyer, Adhiguna Kuncoro, Miguel Ballesteros, and Noah~A Smith.
\newblock Recurrent neural network grammars.
\newblock \emph{arXiv preprint arXiv:1602.07776}, 2016.

\bibitem[El~Hihi \& Bengio(1996)El~Hihi and Bengio]{ElHihi+Bengio-nips8}
Salah El~Hihi and Yoshua Bengio.
\newblock Hierarchical recurrent neural networks for long-term dependencies.
\newblock 1996.
\newblock URL
  \url{http://www.iro.umontreal.ca/~lisa/pointeurs/elhihi_bengio_96.pdf}.

\bibitem[Emami \& Jelinek(2005)Emami and Jelinek]{emami2005neural}
Ahmad Emami and Frederick Jelinek.
\newblock A neural syntactic language model.
\newblock \emph{Machine learning}, 60\penalty0 (1-3):\penalty0 195--227, 2005.

\bibitem[Grave et~al.(2016)Grave, Joulin, and Usunier]{grave2016improving}
Edouard Grave, Armand Joulin, and Nicolas Usunier.
\newblock Improving neural language models with a continuous cache.
\newblock \emph{arXiv preprint arXiv:1612.04426}, 2016.

\bibitem[Ha et~al.(2016)Ha, Dai, and Le]{ha2016hypernetworks}
David Ha, Andrew Dai, and Quoc~V Le.
\newblock Hypernetworks.
\newblock \emph{arXiv preprint arXiv:1609.09106}, 2016.

\bibitem[Hashemi \& Hwa(2016)Hashemi and Hwa]{hashemi2016evaluation}
Homa~B Hashemi and Rebecca Hwa.
\newblock An evaluation of parser robustness for ungrammatical sentences.
\newblock In \emph{EMNLP}, pp.\  1765--1774, 2016.

\bibitem[Hochreiter \& Schmidhuber(1997)Hochreiter and
  Schmidhuber]{hochreiter1997long}
Sepp Hochreiter and J{\"u}rgen Schmidhuber.
\newblock Long short-term memory.
\newblock \emph{Neural computation}, 9\penalty0 (8):\penalty0 1735--1780, 1997.

\bibitem[Inan et~al.(2016)Inan, Khosravi, and Socher]{inan2016tying}
Hakan Inan, Khashayar Khosravi, and Richard Socher.
\newblock Tying word vectors and word classifiers: A loss framework for
  language modeling.
\newblock \emph{arXiv preprint arXiv:1611.01462}, 2016.

\bibitem[Kim et~al.(2016)Kim, Jernite, Sontag, and Rush]{kim2016character}
Yoon Kim, Yacine Jernite, David Sontag, and Alexander~M Rush.
\newblock Character-aware neural language models.
\newblock In \emph{AAAI}, pp.\  2741--2749, 2016.

\bibitem[Klein \& Manning(2002)Klein and Manning]{klein2002generative}
Dan Klein and Christopher~D Manning.
\newblock A generative constituent-context model for improved grammar
  induction.
\newblock In \emph{Proceedings of the 40th Annual Meeting on Association for
  Computational Linguistics}, pp.\  128--135. Association for Computational
  Linguistics, 2002.

\bibitem[Klein \& Manning(2003)Klein and
  Manning]{Klein:2003:AUP:1075096.1075150}
Dan Klein and Christopher~D. Manning.
\newblock Accurate unlexicalized parsing.
\newblock In \emph{Proceedings of the 41st Annual Meeting on Association for
  Computational Linguistics - Volume 1}, ACL '03, pp.\  423--430, Stroudsburg,
  PA, USA, 2003. Association for Computational Linguistics.
\newblock \doi{10.3115/1075096.1075150}.
\newblock URL \url{https://doi.org/10.3115/1075096.1075150}.

\bibitem[Klein \& Manning(2004)Klein and Manning]{klein2004corpus}
Dan Klein and Christopher~D Manning.
\newblock Corpus-based induction of syntactic structure: Models of dependency
  and constituency.
\newblock In \emph{Proceedings of the 42nd Annual Meeting on Association for
  Computational Linguistics}, pp.\  478. Association for Computational
  Linguistics, 2004.

\bibitem[Klein \& Manning(2005)Klein and Manning]{klein2005natural}
Dan Klein and Christopher~D Manning.
\newblock Natural language grammar induction with a generative
  constituent-context model.
\newblock \emph{Pattern recognition}, 38\penalty0 (9):\penalty0 1407--1419,
  2005.

\bibitem[Koutnik et~al.(2014)Koutnik, Greff, Gomez, and
  Schmidhuber]{koutnik2014clockwork}
Jan Koutnik, Klaus Greff, Faustino Gomez, and Juergen Schmidhuber.
\newblock A clockwork rnn.
\newblock In \emph{International Conference on Machine Learning}, pp.\
  1863--1871, 2014.

\bibitem[Krueger \& Memisevic(2015)Krueger and
  Memisevic]{krueger2015regularizing}
David Krueger and Roland Memisevic.
\newblock Regularizing rnns by stabilizing activations.
\newblock \emph{arXiv preprint arXiv:1511.08400}, 2015.

\bibitem[Krueger et~al.(2016)Krueger, Maharaj, Kram{\'a}r, Pezeshki, Ballas,
  Ke, Goyal, Bengio, Larochelle, Courville, et~al.]{krueger2016zoneout}
David Krueger, Tegan Maharaj, J{\'a}nos Kram{\'a}r, Mohammad Pezeshki, Nicolas
  Ballas, Nan~Rosemary Ke, Anirudh Goyal, Yoshua Bengio, Hugo Larochelle, Aaron
  Courville, et~al.
\newblock Zoneout: Regularizing rnns by randomly preserving hidden activations.
\newblock \emph{arXiv preprint arXiv:1606.01305}, 2016.

\bibitem[Kuncoro et~al.(2016)Kuncoro, Ballesteros, Kong, Dyer, Neubig, and
  Smith]{kuncoro2016recurrent}
Adhiguna Kuncoro, Miguel Ballesteros, Lingpeng Kong, Chris Dyer, Graham Neubig,
  and Noah~A Smith.
\newblock What do recurrent neural network grammars learn about syntax?
\newblock \emph{arXiv preprint arXiv:1611.05774}, 2016.

\bibitem[LeCun et~al.(2015)LeCun, Bengio, and Hinton]{lecun2015deep}
Yann LeCun, Yoshua Bengio, and Geoffrey Hinton.
\newblock Deep learning.
\newblock \emph{Nature}, 521\penalty0 (7553):\penalty0 436--444, 2015.

\bibitem[Lin et~al.(1998)Lin, Horne, Tino, and Giles]{lin1998learning}
Tsungnan Lin, Bill~G Horne, Peter Tino, and C~Lee Giles.
\newblock Learning long-term dependencies is not as difficult with narx
  recurrent neural networks.
\newblock Technical report, 1998.

\bibitem[Mahoney(2011)]{mahoney2011large}
Matt Mahoney.
\newblock Large text compression benchmark, 2011.

\bibitem[Marcus et~al.(1993)Marcus, Marcinkiewicz, and
  Santorini]{marcus1993building}
Mitchell~P Marcus, Mary~Ann Marcinkiewicz, and Beatrice Santorini.
\newblock Building a large annotated corpus of english: The penn treebank.
\newblock \emph{Computational linguistics}, 19\penalty0 (2):\penalty0 313--330,
  1993.

\bibitem[Marecek(2016)]{marecek2016twelve}
David Marecek.
\newblock Twelve years of unsupervised dependency parsing.
\newblock In \emph{ITAT}, pp.\  56--62, 2016.

\bibitem[Melis et~al.(2017)Melis, Dyer, and Blunsom]{melis2017state}
G{\'a}bor Melis, Chris Dyer, and Phil Blunsom.
\newblock On the state of the art of evaluation in neural language models.
\newblock \emph{arXiv preprint arXiv:1707.05589}, 2017.

\bibitem[Merity et~al.(2016)Merity, Xiong, Bradbury, and
  Socher]{merity2016pointer}
Stephen Merity, Caiming Xiong, James Bradbury, and Richard Socher.
\newblock Pointer sentinel mixture models.
\newblock \emph{arXiv preprint arXiv:1609.07843}, 2016.

\bibitem[Mikolov \& Zweig(2012)Mikolov and Zweig]{mikolov2012context}
Tomas Mikolov and Geoffrey Zweig.
\newblock Context dependent recurrent neural network language model.
\newblock \emph{SLT}, 12:\penalty0 234--239, 2012.

\bibitem[Mikolov et~al.(2010)Mikolov, Karafi{\'a}t, Burget, Cernock{\`y}, and
  Khudanpur]{mikolov2010recurrent}
Tomas Mikolov, Martin Karafi{\'a}t, Lukas Burget, Jan Cernock{\`y}, and Sanjeev
  Khudanpur.
\newblock Recurrent neural network based language model.
\newblock In \emph{Interspeech}, volume~2, pp.\ ~3, 2010.

\bibitem[Mikolov et~al.(2012)Mikolov, Sutskever, Deoras, Le, Kombrink, and
  Cernocky]{mikolov2012subword}
Tom{\'a}{\v{s}} Mikolov, Ilya Sutskever, Anoop Deoras, Hai-Son Le, Stefan
  Kombrink, and Jan Cernocky.
\newblock Subword language modeling with neural networks.
\newblock \emph{preprint (http://www. fit. vutbr. cz/imikolov/rnnlm/char.
  pdf)}, 2012.

\bibitem[Mikolov et~al.(2013)Mikolov, Chen, Corrado, and
  Dean]{mikolov2013efficient}
Tomas Mikolov, Kai Chen, Greg Corrado, and Jeffrey Dean.
\newblock Efficient estimation of word representations in vector space.
\newblock \emph{arXiv preprint arXiv:1301.3781}, 2013.

\bibitem[Mikolov et~al.(2014)Mikolov, Joulin, Chopra, Mathieu, and
  Ranzato]{mikolov2014learning}
Tomas Mikolov, Armand Joulin, Sumit Chopra, Michael Mathieu, and Marc'Aurelio
  Ranzato.
\newblock Learning longer memory in recurrent neural networks.
\newblock \emph{arXiv preprint arXiv:1412.7753}, 2014.

\bibitem[Press \& Wolf(2017)Press and Wolf]{E17-2025}
Ofir Press and Lior Wolf.
\newblock Using the output embedding to improve language models.
\newblock In \emph{Proceedings of the 15th Conference of the European Chapter
  of the Association for Computational Linguistics: Volume 2, Short Papers},
  pp.\  157--163. Association for Computational Linguistics, 2017.
\newblock URL \url{http://www.aclweb.org/anthology/E17-2025}.

\bibitem[Roark(2001)]{roark2001probabilistic}
Brian Roark.
\newblock Probabilistic top-down parsing and language modeling.
\newblock \emph{Computational linguistics}, 27\penalty0 (2):\penalty0 249--276,
  2001.

\bibitem[Roark \& Hollingshead(2008)Roark and
  Hollingshead]{roark2008classifying}
Brian Roark and Kristy Hollingshead.
\newblock Classifying chart cells for quadratic complexity context-free
  inference.
\newblock In \emph{Proceedings of the 22nd International Conference on
  Computational Linguistics-Volume 1}, pp.\  745--751. Association for
  Computational Linguistics, 2008.

\bibitem[Sandra \& Taft(1994)Sandra and Taft]{sandra1994morphological}
Dominiek Sandra and Marcus Taft.
\newblock \emph{Morphological structure, lexical representation and lexical
  access}.
\newblock Taylor \& Francis, 1994.

\bibitem[Schmidhuber(2015)]{schmidhuber2015deep}
J{\"u}rgen Schmidhuber.
\newblock Deep learning in neural networks: An overview.
\newblock \emph{Neural networks}, 61:\penalty0 85--117, 2015.

\bibitem[Schmidhuber(1991)]{Schmidhuber91neuralsequence}
Jürgen Schmidhuber.
\newblock Neural sequence chunkers.
\newblock Technical report, 1991.

\bibitem[Socher et~al.(2010)Socher, Manning, and Ng]{socher2010learning}
Richard Socher, Christopher~D Manning, and Andrew~Y Ng.
\newblock Learning continuous phrase representations and syntactic parsing with
  recursive neural networks.
\newblock In \emph{Proceedings of the NIPS-2010 Deep Learning and Unsupervised
  Feature Learning Workshop}, pp.\  1--9, 2010.

\bibitem[Socher et~al.(2013)Socher, Perelygin, Wu, Chuang, Manning, Ng, and
  Potts]{socher2013recursive}
Richard Socher, Alex Perelygin, Jean Wu, Jason Chuang, Christopher~D Manning,
  Andrew Ng, and Christopher Potts.
\newblock Recursive deep models for semantic compositionality over a sentiment
  treebank.
\newblock In \emph{Proceedings of the 2013 conference on empirical methods in
  natural language processing}, pp.\  1631--1642, 2013.

\bibitem[Solan et~al.(2003)Solan, Ruppin, Horn, and
  Edelman]{solan2003automatic}
Zach Solan, Eytan Ruppin, David Horn, and Shimon Edelman.
\newblock Automatic acquisition and efficient representation of syntactic
  structures.
\newblock In \emph{Advances in Neural Information Processing Systems}, pp.\
  107--114, 2003.

\bibitem[Sukhbaatar et~al.(2015)Sukhbaatar, Weston, Fergus,
  et~al.]{sukhbaatar2015end}
Sainbayar Sukhbaatar, Jason Weston, Rob Fergus, et~al.
\newblock End-to-end memory networks.
\newblock In \emph{Advances in neural information processing systems}, pp.\
  2440--2448, 2015.

\bibitem[Tai et~al.(2015)Tai, Socher, and Manning]{tai2015improved}
Kai~Sheng Tai, Richard Socher, and Christopher~D Manning.
\newblock Improved semantic representations from tree-structured long
  short-term memory networks.
\newblock \emph{arXiv preprint arXiv:1503.00075}, 2015.

\bibitem[Titov \& Henderson(2010)Titov and Henderson]{titov2010latent}
Ivan Titov and James Henderson.
\newblock A latent variable model for generative dependency parsing.
\newblock In \emph{Trends in Parsing Technology}, pp.\  35--55. Springer, 2010.

\bibitem[Williams et~al.(2017)Williams, Drozdov, and
  Bowman]{williams2017learning}
Adina Williams, Andrew Drozdov, and Samuel~R Bowman.
\newblock Learning to parse from a semantic objective: It works. is it syntax?
\newblock \emph{arXiv preprint arXiv:1709.01121}, 2017.

\bibitem[Wu et~al.(2017)Wu, Zhang, Yang, Li, and Zhou]{wu2017sequence}
Shuangzhi Wu, Dongdong Zhang, Nan Yang, Mu~Li, and Ming Zhou.
\newblock Sequence-to-dependency neural machine translation.
\newblock In \emph{Proceedings of the 55th Annual Meeting of the Association
  for Computational Linguistics (Volume 1: Long Papers)}, volume~1, pp.\
  698--707, 2017.

\bibitem[Wu et~al.(2016)Wu, Zhang, Zhang, Bengio, and
  Salakhutdinov]{wu2016multiplicative}
Yuhuai Wu, Saizheng Zhang, Ying Zhang, Yoshua Bengio, and Ruslan~R
  Salakhutdinov.
\newblock On multiplicative integration with recurrent neural networks.
\newblock In \emph{Advances in Neural Information Processing Systems}, pp.\
  2856--2864, 2016.

\bibitem[Zaremba et~al.(2014)Zaremba, Sutskever, and
  Vinyals]{zaremba2014recurrent}
Wojciech Zaremba, Ilya Sutskever, and Oriol Vinyals.
\newblock Recurrent neural network regularization.
\newblock \emph{arXiv preprint arXiv:1409.2329}, 2014.

\bibitem[Zhang et~al.(2015)Zhang, Lu, and Lapata]{zhang2015top}
Xingxing Zhang, Liang Lu, and Mirella Lapata.
\newblock Top-down tree long short-term memory networks.
\newblock \emph{arXiv preprint arXiv:1511.00060}, 2015.

\bibitem[Zhou et~al.(2017)Zhou, Luo, Cao, Xiao, Lin, Chen, and
  He]{zhou2017generative}
Ganbin Zhou, Ping Luo, Rongyu Cao, Yijun Xiao, Fen Lin, Bo~Chen, and Qing He.
\newblock Generative neural machine for tree structures.
\newblock \emph{arXiv preprint arXiv:1705.00321}, 2017.

\bibitem[Zilly et~al.(2016)Zilly, Srivastava, Koutn{\'\i}k, and
  Schmidhuber]{zilly2016recurrent}
Julian~Georg Zilly, Rupesh~Kumar Srivastava, Jan Koutn{\'\i}k, and J{\"u}rgen
  Schmidhuber.
\newblock Recurrent highway networks.
\newblock \emph{arXiv preprint arXiv:1607.03474}, 2016.

\bibitem[Zoph \& Le(2016)Zoph and Le]{zoph2016neural}
Barret Zoph and Quoc~V Le.
\newblock Neural architecture search with reinforcement learning.
\newblock \emph{arXiv preprint arXiv:1611.01578}, 2016.

\end{thebibliography}
\bibliographystyle{iclr2017_conference}

\section*{APPENDIX}
\appendix

\section{Inferred Tree Structure}
\label{appendix_tree_fig}
\begin{figure}[h!]
    \centering
    \setlength{\leftskip}{-30pt}
    \begin{subfigure}[b]{0.49\textwidth}
        \includegraphics[width=2.3\textwidth, angle=90]{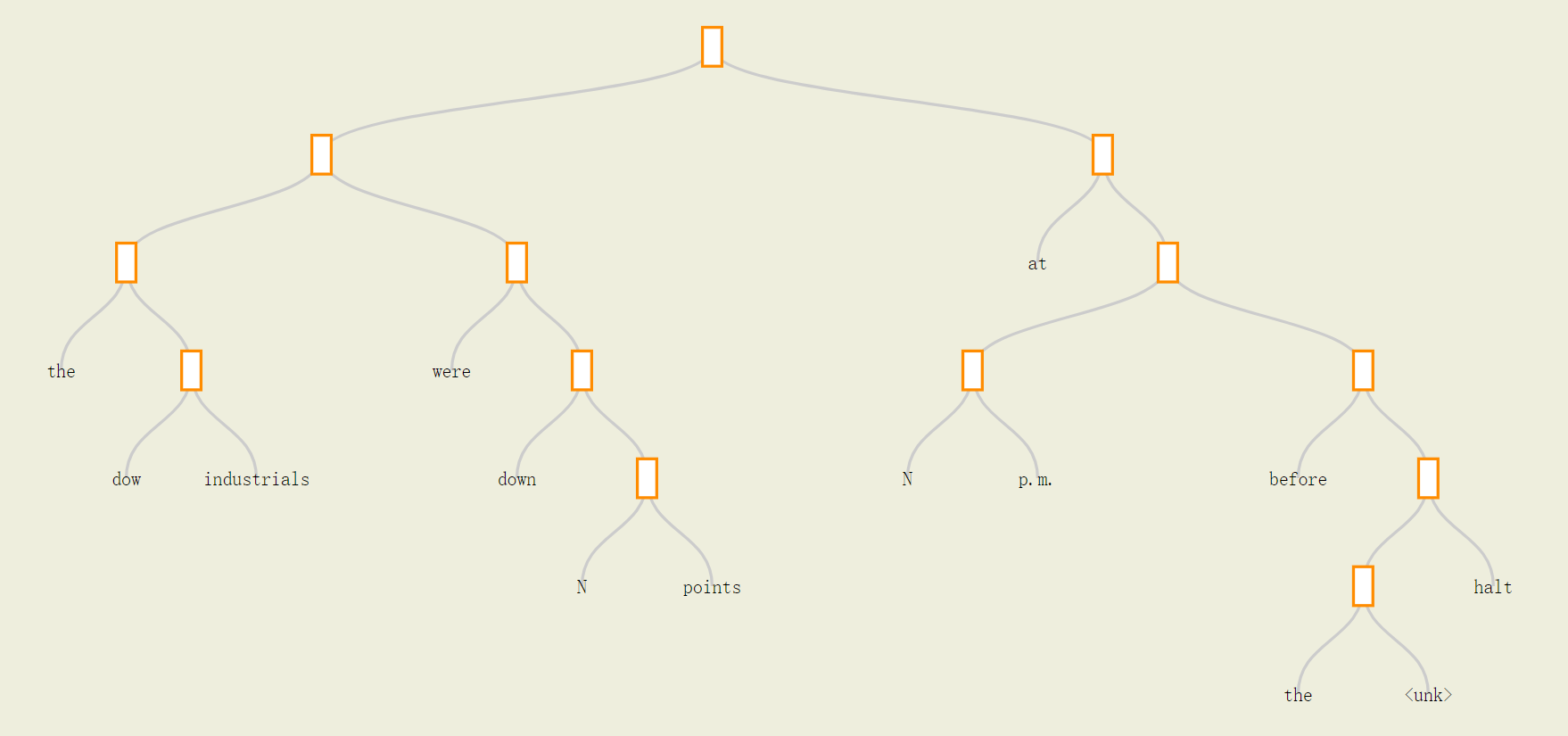}
    \end{subfigure}
    ~ 
    \begin{subfigure}[b]{0.49\textwidth}
        \includegraphics[width=2.3\textwidth, angle=90]{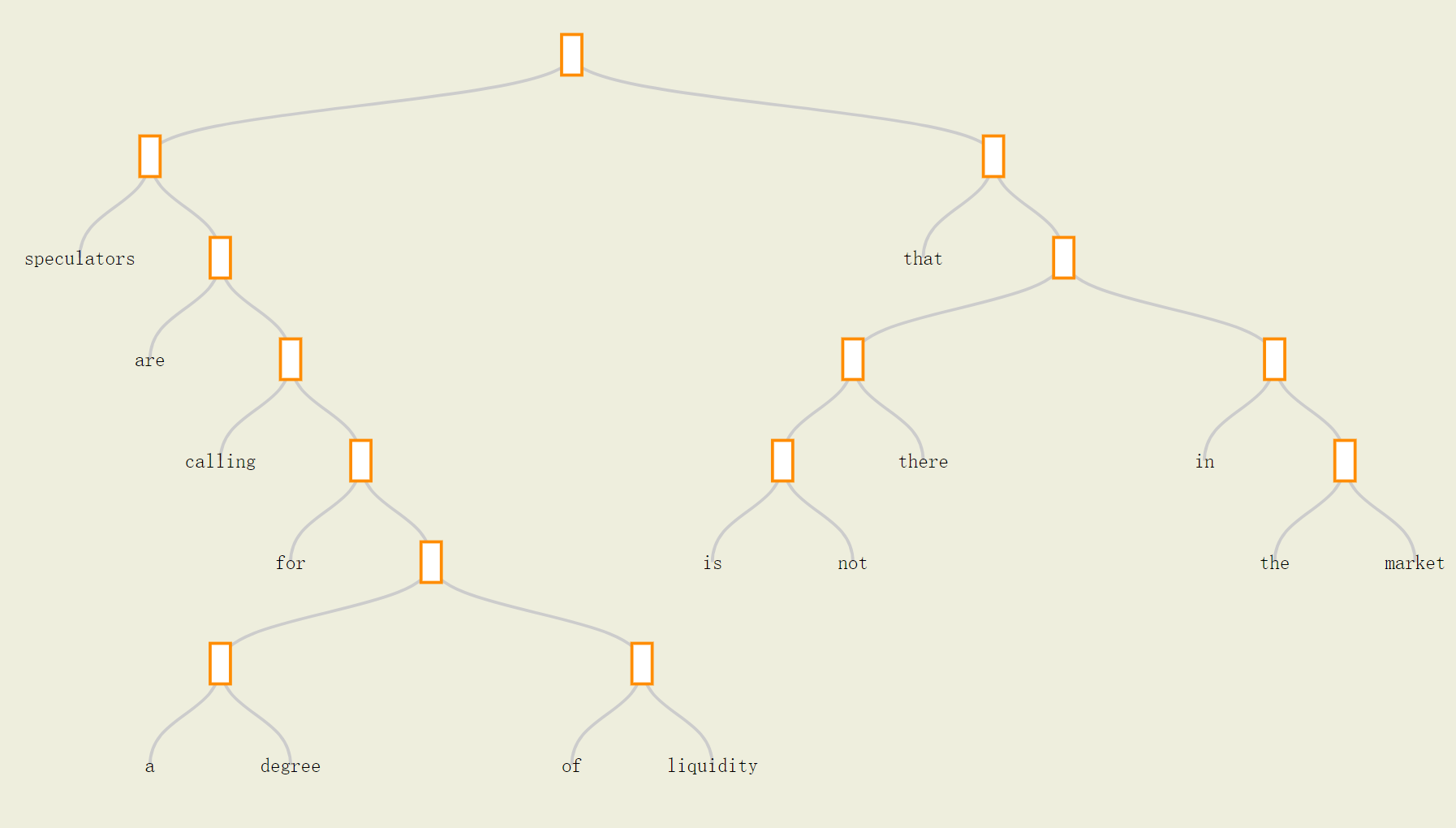}
    \end{subfigure}
    \caption{Syntactic structures of two different sentences inferred from $\{d_i\}$ given by Parsing Network.}
\end{figure}

The tree structure is inferred from the syntactic distances yielded by the Parsing Network. We first sort the $d_i$'s in decreasing order. 
For the first $d_i$ in the sorted sequence, we separate sentence into constituents $((x_{<i}), (x_{i}, (x_{>i})))$. Then we separately repeat this operation for constituents $(x_{<i})$ and $(x_{>i})$. Until the constituent contains only one word.

\section{Modeling Local Structure} \label{appendix_MLS}
In this section we give a probabilistic view on how to model the local structure of language. 
Given the nature of language, sparse connectivity can be enforced as a prior on how to improve generalization and interpretability of the model.
% Hence, we propose to introduce a latent variable $\gamma_{t+1}$ that determines the structure of a sentence, and thus the graph structure (i.e. conditional dependencies) of any word in the sentence.

% Specifically, we define 

% \begin{equation}
% p(x_{t+1}|x_0, ..., x_t) = \sum_{t'=0}^t p(x_{t+1}|x_{t'}, ..., x_t)p(\gamma_{t+1}=t'|x_0, ...,x_t)
% \end{equation}

At time step $t$, $p(l_{t}|x_0, ...,x_t)$ represents the probability of choosing one out of $t$ possible local structures that defines the conditional dependencies. 
If $l_{t}=t'$, it means $x_{t}$ depends on all the previous hidden state from $m_{t'}$ to $m_{t}$ ($t'\leq t$). 

A particularly flexible option for modeling $p(l_{t}|x_0, ..., x_t)$ is the Dirichlet Process, since being non-parametric allows us to attend on as many words as there are in a sentence; 
i.e. number of possible structures (mixture components) grows with the length of the sentence.
As a result, we can write the probability of $l_{t+1}=t'$ as a consequence of the stick breaking process \footnote{Note that the index is in decreasing order.}:
\begin{equation}
p(l_{t}=t'|x_0, ...,x_t) = (1-\alpha_{t'}^{t})\prod_{j=t'+1}^{t-1}\alpha_j^{t}
\end{equation}
for $1\leq t'< t-1$, and 
\begin{equation}
p(l_{t}=t-1|x_0, ...,x_t) = (1-\alpha_{t-1}^{t}) ;\qquad p(l_{t}=0|x_0, ...,x_t) = \prod_{j=1}^{t-1}\alpha_j^{t}
\end{equation}
where $\alpha_j = 1 - \beta_j$ and $\beta_j$ is a sample from a Beta distribution. 
Once we sample $l_{t}$ from the process, the connectivity is realized by a element-wise multiplication of an attention weight vector with a masking vector $g_{t}$ defined in Eq. \ref{eq_gt}.
In this way, $x_t$ becomes functionally independent of all $x_s$ for all $s<l_{t}$.
The expectation of this operation is the CDF of the probability of $l$, since 
%Aaron: in the above do you mean this or do you mean CDF of $l$?

%\begin{equation}
%\begin{split}
%\mathbf{E}_{l_{t+1}}[v_{t+1}^{\{i\}}]
%&=(1-\alpha_{t'})\prod_{j=t'+1}\alpha_j +
%(1-\alpha_{t'+1})\prod_{j=t'+2}\alpha_j + %... + 
%(1-\alpha_{t}) \\
%&= \sum_{k=t'}^t %(1-\alpha_{k})\prod_{j=k+1}^t\alpha_j = %\mathbf{P}(l_{t+1}\geq t')
%\end{split}
%\end{equation}

\begin{equation}
\begin{split}
\mathbf{E}_{l_{t}}[g_{t}^{\{t'\}}]
&=\prod_{j=1}\alpha_j^{t} + (1-\alpha_1^{t})\prod_{j=2}\alpha_j^{t} + ... + (1-\alpha_{t'}^{t})\prod_{j=t'+1}\alpha_j^{t} \\
&= \sum_{k=0}^{t'} p(l_{t}=k|x_0, ...,x_t) = \mathbf{P}(l_{t}\leq t')
\end{split}
\end{equation}

By telescopic cancellation, the CDF can be expressed in a succinct way: 

\begin{equation}
\mathbf{P}(l_{t}\leq t') = \prod_{j=t'+1}^{t-1}\alpha_j^{t}
\end{equation}

for $t'<t$, and $\mathbf{P}(l_{t}\leq t) = 1$.
However, being Bayesian nonparametric and assuming a latent variable model require approximate inference. 
%Aaron: what the problem with approximate inference? This is a strange transition. Are your relaxations implementing a well defined approximation or is this a heuristic. Either way is ok, but we should probably specify
Hence, we have the following relaxations

\begin{enumerate}
	\item First, we relax the assumption and parameterize $\alpha_j^{t}$ as a deterministic function depending on all the previous words, which we will describe in the next section. 
	\item We replace the discrete decision on the graph structure with a soft attention mechanism, by multiplying attention weight with the multiplicative gate:
    \begin{equation}
      g_i^{t} = \prod_{j=i+1}^t \alpha_j^{t}
    \end{equation}
\end{enumerate}

With these relaxations, Eq. (\ref{eq_cond}) can be approximated by using a soft gating vector to update the hidden state $h$ and the predictive function $f$. This approximation is reasonable since the gate is the expected value of the discrete masking operation described above.

\section{No Partial Overlapping in Dependency Ranges} \label{appendix_overlap}
In this appendix, we show that having no partial overlapping in dependency ranges is an essential property for recovering a valid tree structure, and PRPN can provide a binary version of $g_i^{t}$, that have this property.

The masking vector $g_i^{t}$ introduced in Section \ref{MLS} determines the range of dependency, i.e., for the word $x_{t}$ we have $g_i^{t}=1$ for all $l_t \le i < t$. All the words fall into the range $l_t \le i < t$ is considered as $x_t$'s sibling or offspring of its sibling. If the dependency ranges of two words are disjoint with each other, that means the two words belong to two different subtrees. If one range contains another, that means the one with smaller range is a sibling, or is an offspring of a sibling of the other word. However, if they partially overlaps, they can't form a valid tree. 
%In that respect, we have to parameterize $\alpha_j^t$ carefully so that the structure given by $g_i^t$'s approximates a valid tree. 

While Eq.\ref{gate_equal_multialpha} and Eq.\ref{soft_alpha} provide a soft version of dependency range, we can recover a binary version by setting $\tau$ in Eq.\ref{soft_alpha} to $+\infty$. The binary version of $\alpha_j^t$ corresponding to Eq. \ref{soft_alpha} becomes:
\begin{equation}  \label{hard_alpha}
\alpha_j^t = \frac{\mathrm{sign} \left( d_t - d_{j+1} \right) + 1}{2}
\end{equation}
which is basically the sign of comparing $d_t$ and $d_{j+1}$, scaled to the range of 0 and 1. Then for each of its previous token the \emph{gate} value $g_i^t$ can be computed through Eq.\ref{gate_equal_multialpha}. 

Now for a certain $x_t$, we have 
\begin{equation}
g_i^t= \begin{cases} 1,\quad t' \le i < t \\  0,\quad 0 \le i < t' \end{cases}
\end{equation}
where
\begin{equation}
t'= \max{i}, \quad s.t. \quad d_i > d_t
\end{equation}
Now all the words that fall into the range $t' \le i < t$ are considered as either sibling of $x_t$, or offspring of a sibling of $x_t$ (Figure \ref{fig_conv}). The essential point here is that, under this parameterization, the dependency range of any two tokens won't partially overlap. Here we provide a terse proof:

\begin{proof}
Let's assume that the dependency range of $x_v$ and $x_n$ partially overlaps. We should have 
$g_i^u=1$ for $u \le i < v$ and $g_i^n=1$ for $m \le i < n$. Without losing generality, we assume 
$u<m<v<n$ so that the two dependency ranges overlap in the range $[m, v]$. 
\begin{enumerate}
\item For $x_v$, we have $\alpha_i^v=1$ for all $u \le i < v$. According to Eq. \ref{soft_alpha} and \ref{gate_equal_multialpha}, we have $d_i < d_v$ for all $u \le i < v$. Since $u<m$, we have $d_m < d_v$.
\item Similarly, for $x_n$, we have $d_i < d_n$ for all $m \le i < n$. Since $m<v$, we have $d_v < d_n$. On the other hand, since the range stops at $m$, we should also have $d_m > d_n$. Thus $d_m > d_v$.
\end{enumerate}
Items 1 and 2 are contradictory, so the dependency ranges of $x_v$ and $x_n$ won't partially overlap.
\end{proof}

\section{Properties and Intuitions of $g_i^t$ and $d_i$} 
\label{gd_properties}
First, for any fixed $t$, $g_i^t$ is monotonic in $i$. This ensures that $g_i^t$ still provides soft truncation to define a dependency range. 

The second property comes from $\tau$. The hyperparameter $\tau$ has an interesting effect on the tree structure: if it is set to 0, then for all $t$, the gates $g_i^t$ will be open to all of $e_t$'s predecessors, which will result in a flat tree where all tokens are direct children of the root node; as $\tau$ becomes larger, the number of levels of hierarchy in the tree increases. As it approaches $+\inf$, the $\mathrm{hardtanh}(\cdot)$ becomes $\mathrm{sign}(\cdot)$ and the dependency ranges form a valid tree. Note that, due to the linear part of the gating mechanism, which benefits training, when $\tau$ has a value in between the two extremes the truncation range for each token may overlap. That may sometimes result in vagueness in some part of the inferred tree. To eliminate this vagueness and ensure a valid tree, at test time we use $\tau=+\inf$.

Under this framework, the values of syntactic distance have more intuitive meanings. If two adjacent words are siblings of each other, the syntactic distance should approximate zero; otherwise, if they belong to different subtrees, they should have a larger syntactic distance. In the extreme case, the syntactic distance approaches 1 if the two words have no subtree in common. In Figure \ref{fig_conv} we show the syntactic distances for each adjacent token pair which results in the tree shown in Figure \ref{fig_tree}.

\section{Baseline Methods for Unsupervised Constituency Parsing} \label{appendix_parser_baseline}
Our model is compared with the same baseline methods as in \citep{klein2005natural}. \emph{RANDOM} chooses a binary tree uniformly at random from the set of binary trees. This is the unsupervised baseline. \emph{LBRANCH} and \emph{RBRANCH} choose the completely left- and right-branching structures, respectively. \emph{RBRANCH} is a frequently used baseline for supervised parsing, but it should be stressed that it encodes a significant fact about English structure, and an induction system need not beat it to claim a degree of success. \emph{UPPER BOUND} is the upper bound on how well a binary system can do against the Treebank sentences. Because the Treebank sentences are generally more flat than binary, limiting the maximum precision which can be attained, since additional brackets added to provide a binary tree will be counted as wrong.

We also compared our model with other unsupervised constituency parsing methods. \emph{DEP-PCFG} is dependency-structured PCFG \citep{carroll1992two}. \emph{CCM} is constituent-context model \citep{klein2002generative}. \emph{DMV} is an unsupervised dependency parsing model. \emph{DMV+CCM} is a combined model that jointly learn both constituency and dependency parser \citep{klein2004corpus}.

\appendix

\end{document}